# Decision-making at Unsignalized Intersection for Autonomous Vehicles: Left-turn Maneuver with Deep Reinforcement Learning

Feng Wang, Dongjie Shi, Teng Liu, Xiaolin Tang

***Abstract*—Decision-making module enables autonomous vehicles to reach appropriate maneuvers in the complex urban environments, especially the intersection situations. This work proposes a deep reinforcement learning (DRL) based left-turn decision-making framework at unsignalized intersection for autonomous vehicles. The objective of the studied automated vehicle is to make an efficient and safe left-turn maneuver at a four-way unsignalized intersection. The exploited DRL methods include deep Q-learning (DQL) and double DQL. Simulation results indicate that the presented decision-making strategy could efficaciously reduce the collision rate and improve transport efficiency. This work also reveals that the constructed left-turn control structure has a great potential to be applied in real-time.**

## NOMENCLATURE

| | |
|---|---|
| DRL | Deep Reinforcement Learning |
| DQL | Deep Q-learning |
| RL | Reinforcement Learning |
| DDPG | Deep Deterministic Policy Gradient |
| DQN | Deep Q-network |
| EAV | Ego Automated Vehicle |
| IDM | Intelligent Driver Model |
| ACC | Adaptive Cruise Control |
| MDP | Markov Decision Process |
| DDQL | Double DQL |

## I. INTRODUCTION

URBAN transportation environment is complicated and uncertain, which motivates academic researchers and vehicle manufacturers to develop advanced technologies for self-driving [1-2]. The intersection is a representative driving scenario in an urban environment and it contains multiple participants, such as pedestrians, motor vehicles, traffic lights, and nonmotor vehicles [3]. In order to achieve autonomous driving with high automation, the self-driving cars should dispose of the unsignaled intersections appropriately [4-5]. However, some challenges still remain for left-turn maneuvers in regard to the unsignaled intersections [6].

Decision-making controller is completely important for autonomous vehicles. This module enables the self-driving cars to execute right navigated behaviors to accomplish a given driving mission [7-8]. In the unsignaled intersection situation, decision-making indicates running through (left-turn, right-turn, or go straight) the intersection according to the criteria of right-of-way. For left-turn maneuver, the automated vehicles need to yield the pedestrians and straight vehicles with safety and efficiency objectives. Without the guidance of traffic lights, the self-driving cars should adapt to the uncertain intentions of the surrounding vehicles [9-11].

In the literature of decision-making at intersection situations, the methods have two categories, rule-based approaches and learning-based ones [12]. Rule-based decision-making policies are often constructed by human knowledge and engineering experiences. For example, the authors in [13] proposed a situation hypotheses selection system to decide the behaviors of the studied vehicles. The regression methods are employed to assess the risk of each decision choice. In Ref. [14], Zhao et al. presented a machine understandable ontology based knowledge decision-making system by combining the traffic regulations and maps. This system could plan the route by building a collision warning signal to avoid a crash. Furthermore, the work in [15] established a decentralized coordination approach to handle the decision-making problem. The derived control policy includes the model-based heuristics to reach low complexity solutions and fast online implementation. However, the performance of rule-based techniques really depends on the modeling accuracy and human builder. In addition, the rule-based method does not consider the uncertain intention of the surrounding agents, so the robustness of the method is poor in unknown cases[16].

Reinforcement learning (RL) and deep reinforcement learning (DRL) exhibit great potential in resolving sequential decision-making problems [17-20]. Many attempts have conducted to apply RL or DRL to address the decision-making

This work was supported by Foundation of State Key Laboratory of Automotive Simulation and Control (20171108) and QAII Joint Fund Project 16-7-1-4-JCH. (Corresponding authors: X. Tang, F. Wang)

F. Wang and D. Shi are with School of Economics and Business Administration, Chongqing University, Chongqing, 400044, China. (email: wangfeng2008@cqu.edu.cn, shidongjie@cqu.edu.cn)

T. Liu is with Clinical Research Center (CRC), Clinical Pathology Center (CPC), Chongqing University Three Gorges Hospital, Chongqing University, Wanzhou, Chongqing, China. (email: tengliu17@ gmail.com)

X. Tang is with College of Mechanical and Vehicle Engineering, Chongqing University, Chongqing, 400044, China. (email: tangxiaolin6@126.com)

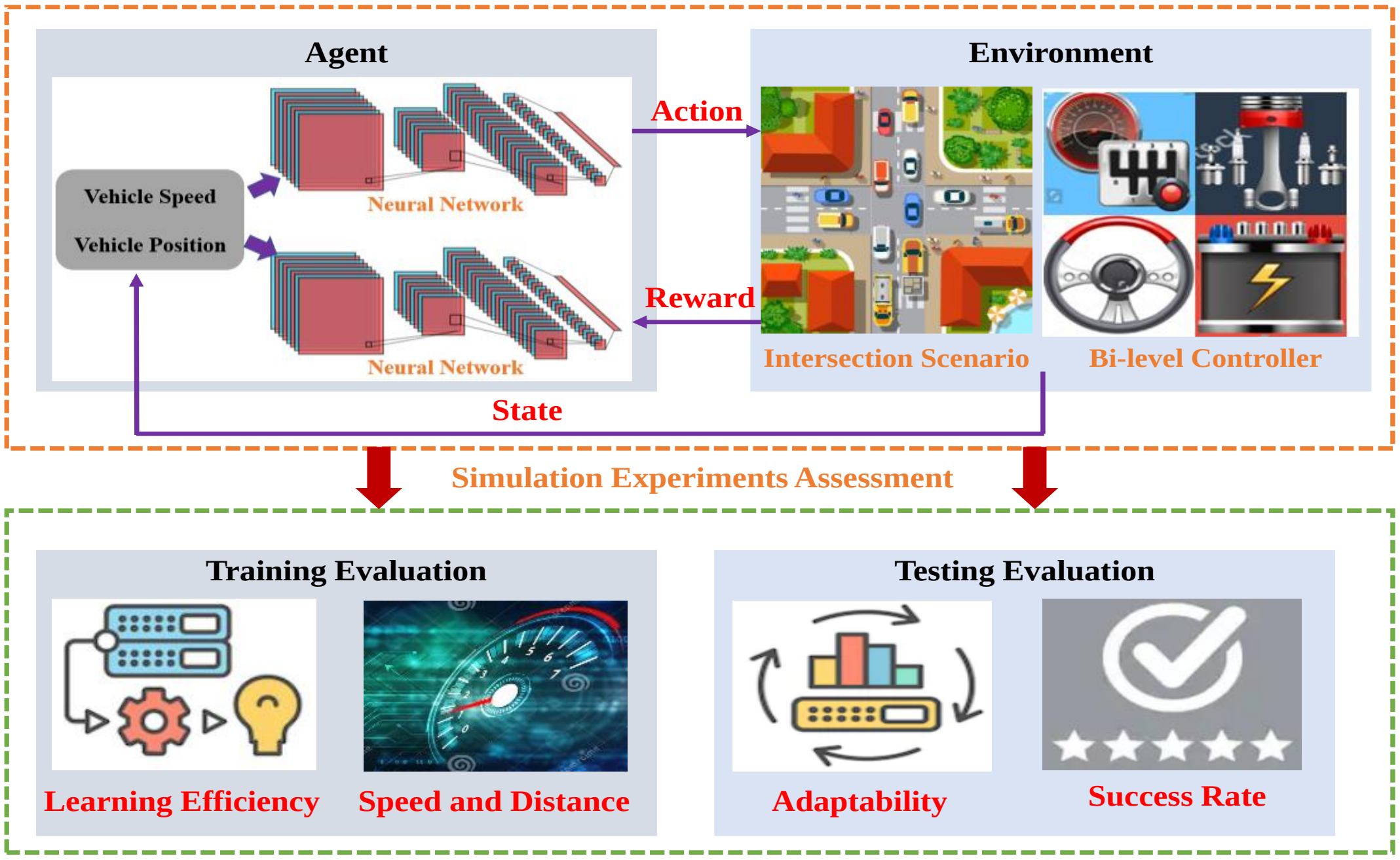

Fig. 1. Control architecture of DRL for decision-making and its evaluation concept for autonomous vehicles

issues at intersections for autonomous driving. For example, Zhou et al. [21] utilized a deep deterministic policy gradient (DDPG) algorithm to exploit an efficient driving strategy for connected and automated vehicles at signalized intersections. The proposed controller is proven to work well under different traffic demands and traffic light cycles. The authors in [22] used the critical turning point idea into a hierarchical decision-making framework. The resulted control policy enables human-friendly autonomous driving and leads to high computational efficiency. Furthermore, Ref. [23-24] explored the deep Q-network (DQN)-based navigating policies at the intersection for self-driving cars. The underlying risk for collision is capable of being predicted and averted, and the task completion time and the goal success rate is able to be improved. However, the suitable approximation of the Q-function in DRL approaches needs to be further discussed.

In this work, a decision-making controller based on deep Q-learning (DQL) is presented for autonomous vehicles at unsignaled intersections, as displayed in Fig. 1. First, the studied driving scenario is established as a four-way intersection, in which the ego automated vehicle (EAV) would like to perform a left turn behavior. The modeling of the EAV and its nearby vehicles are then constructed. Then, the decision-making problem is transformed into a control optimization problem in the RL perspective. The DQL and double DQL algorithms are applied to search the optimal decision-making strategy for EAV. Finally, three simulation experiments are arranged to estimate the effectiveness of the proposed method. The collision rate and success rate are discussed and analyzed to evaluate the relevant decision-making policy.

The original contributions of this paper are illuminated as follows: 1) a DRL-enhanced decision-making controller is proposed to deal with the uncertainties in the unsignaled intersection situation; 2) two DRL algorithms, named DQL and double DQL are compared and assessed in the decision-making problem for self-driving vehicles; 3) multiple evaluation norms are adopted to analyze the optimality and robustness of the derived decision-making strategy. This article is one attempt to discuss the efficiency and safety of decision-making policy at unsignaled intersections for autonomous vehicles.

The following organization of this work is depicted as follows: Section II gives the research driving scenario and vehicle modeling. The RL framework and realization of the DRL algorithms are displayed in Section III. The discussion of the related simulation results is shown in Section IV. The conclusion and future work are presented in Section V.

## II. Unsignaled Intersection and Vehicle Movements

This section mimics the driving conditions for self-driving cars at unsignalized intersections. A four-way intersection scenario is constructed firstly, where the ego automated vehicle (EAV) wants to make a left-turn decision. The modeling of the vehicle movements of the EAV and its nearby vehicles are then given. Finally, the driving behaviors controller is formulated to manage the driving intentions of the surrounding vehicles.

### A. *Driving Scenario of Intersection*

As shown in Fig. 2, an EAV usually has three choices according to the driving task, and they are left-turn, right-turn, and go straight. In the four-way intersection environment, all the vehicles should follow the regulation of right-of-way. The right-of-way indicates which vehicle has the priority to drive through the intersection preferentially. The right-of-way at the unsignaled intersection has four levels. They are Level 4: the horizontal straight moving and right-turn; Level 3: the vertical straight moving and right-turn; Level 2: the horizontal left-turn and Level 1: the vertical left-turn.

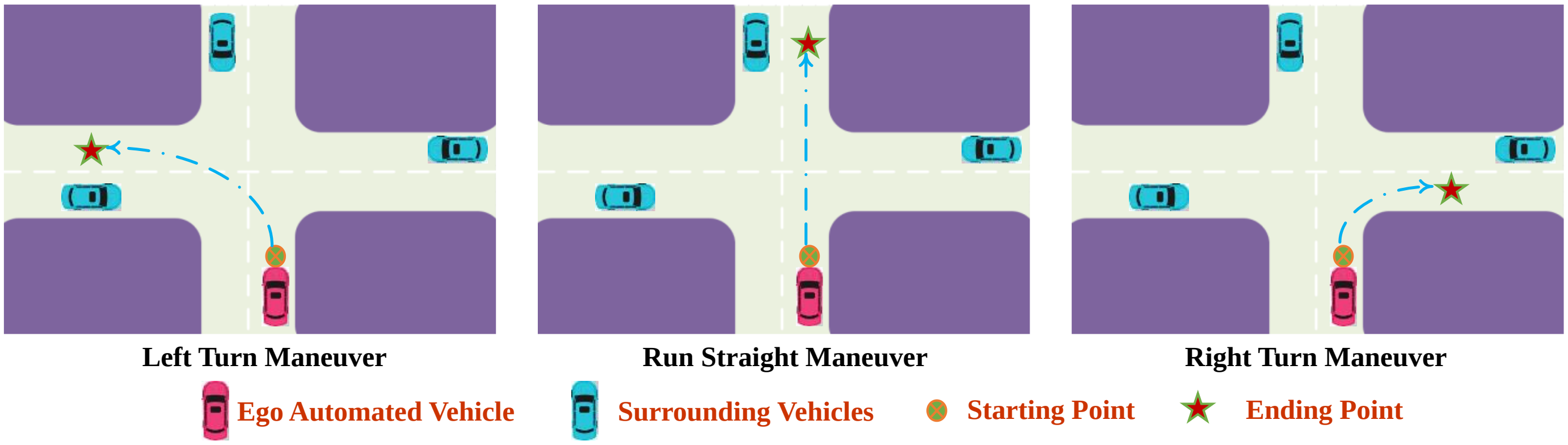

Fig. 2. Decision-making problems at the unsignaled intersection for autonomous vehicles.

Since the vertical left-turn has the lowest priority, this work discusses the left-turn maneuver for the EAV. It means the EAV is necessary to yield other surrounding vehicles and then execute a left-turn. To avoid a crash, the EAV needs to slow down or stop to wait for an opportunity. Hence, efficiency and safety can be regarded as the evaluation criteria for the decision-making policy at the unsignaled intersection. The goal of the EAV is to run as fast as possible from the starting point to the ending point without a collision.

From Fig. 2, the pink vehicle represents the EAV, and other blue vehicles are the surrounding vehicles. The driving intentions of the surrounding vehicles are unknown to the EAV, and thus it should figure out the intentions and make the right decision. The number of the surrounding vehicle is $N$, and the position, speed destination, and driving behaviors of these surrounding vehicles are all random determined. This feature increases the uncertainties of this driving situation. The EAV needs to learn how to manipulate a suitable decision-making strategy.

As for a pivotal concern, the surrounding vehicle may crash its neighbors (another surrounding vehicle) with random characteristics. To evade this problem, a simplistic behaviour is adopted [25]. Each vehicle could forecast the future positions of its neighbours over a three-seconds horizon. Based on the right-of-way, the yielding vehicle should brake to avoid a collision. The longitudinal and lateral movements of the surrounding vehicles are managed by a hierarchical structure, which will be introduced in the next subsection.

Without loss of generality, the default setting of the unsignaled intersection is determined as follows: the amount of surrounding vehicle is $N$=15, and the considered behavior of the EAV is vertical left-turn. The duration of this situation is 15 seconds. The simulation frequency is 20 Hz, and the sampling time is 1 second (the EAV selects control action every one second). The starting and ending points of the EAV are the same in each trial.

## B. *Vehicle Movements*

A hierarchical control framework is applied to regulate the behaviors of the surrounding vehicles and EAV. For the EAV, the high-level is the proposed DRL-enabled decision-making policy to manage the longitudinal behavior, and the low-level is the tracking controller for lateral control. For the surrounding vehicles, the high-level is an intelligent driver model (IDM) [26], and the low-level is also the tracking controller. These two levels are executed at the same time. The hierarchical control framework for the surrounding vehicle is called the reference model in the following content.

The lateral controller involves two aspects of controls, the position control and heading control. The position control is stated as follows:

$$v_l^{ex} = -K_{p,l} d_l \tag{1}$$

$$\Delta\psi = \arcsin\left(\frac{v_l^{ex}}{v}\right) \tag{2}$$

where $v^{ex}{}_l$ is the target lateral speed, $K_{p,l}$ is the position control gain and $d_l$ is the lateral distance between the vehicle and the center-line of the target lane. $\psi$ is the heading angle, and $v$ is the speed. The heading control is computed by a proportional-derivative controller as:

$$\psi^{ex} = \psi_L + \Delta\psi \tag{3}$$

$$\dot{\psi} = K_{p,\psi}(\psi^{ex} - \psi) \tag{4}$$

$$\delta = \arcsin\left(\frac{1}{2}\frac{l_r}{v}\dot{\psi}\right) \tag{5}$$

where $\psi_L$ is the lane heading, $\psi^{ex}$ is the expected heading angle, and $K_{p,\psi}$ is the heading control gain. $\delta$ is the steering angle and $l_r$ is the distance between the center of the front wheel and the center of gravity.

The longitudinal controller is realized by IDM to control the longitudinal acceleration. The IDM is often applied in automated vehicles for adaptive cruise control (ACC) [27]. The relevant acceleration is determined as follows:

$$a = a_{\max} \cdot \left(1 - \left(\frac{v}{v_d}\right)^{\theta} - \left(\frac{s_r}{\Delta s}\right)^2\right) \tag{6}$$

where $a_{\max}$ is the maximum acceleration. $s_d$ and $v_d$ are the demand distance and speed. $\triangle s$ is the interval between the studied vehicle and the leading vehicle, and $\theta$ is the constant acceleration parameter. In IDM, the demand distance is further computed as:

$$s_d = s_0 + T \cdot v + \frac{v \cdot \Delta v}{2\sqrt{a_{\max} \cdot b}} \tag{7}$$

where $s_0$ is the minimum relative distance between two vehicles on the same lane, and $T$ is the desired time interval for the safety objective. $\triangle v$ is the relative speed gap between the EAV and its front one, and $b$ is the value of deceleration according to the comfortable purpose. The parameters of the IDM in this work is depicted in Table I.

TABLE I
MAIN PARAMETERS IN IDM

| Keyword | Value | Unit |
|---|---|---|
| Maximum acceleration $\boldsymbol{a_{max}}$ | 6 | m/s$^2$ |
| Acceleration argument $\boldsymbol{\theta}$ | 4 | / |
| Desired time gap $\boldsymbol{T}$ | 1.5 | s |
| Comfortable deceleration rate $\boldsymbol{b}$ | -5 | m/s$^2$ |
| Minimum relative distance $\boldsymbol{s_0}$ | 10 | m |

## III. Deep Reinforcement Learning Approaches

In this section, we introduced the implementation of the DRL methods. The Markov decision processes (MDPs) and parameters in RL are illustrated first. Then, the Deep Q-learning (DQL) and its variation named double DQL are formulated. Finally, the specification of the decision-making problem is constructed, in which the special state variables, control actions, reward function, and transition model are given.

### A. Definition in RL

The decision-making problem of autonomous vehicles at intersections is a classical sequential decision making under uncertainty and it can be expressed as one MDP. MDPs are problems of learning from agent-environment interaction to achieve a goal [28]. The interaction at each time-step $t$ is that 1) the agent chooses an action $a_t$ to execute at current state $s_t$, 2) the environment transits to a new state $s_{t+1}$ and presents a reward $r_t$ to the agent dependent only on the preceding state and action. All states in MDPs have Markov property and the state-transition probabilities are $P(s_{t+1}) = P_r\,(s_{t+1}|s_t, a_t)$.

In RL, the goal of the agent is to learn an optimal policy to maximize the expected discounted return [29]. The policy $\pi$ maps state to the probabilities of choosing each action. The return $G_t$ is defined as the sum of the discounted future rewards at time $t$:

$$G_t = \sum_{k=t}^{T} \gamma^{k-t} r_k \tag{8}$$

where $\gamma$ is a discounted factor, $T$ is the time-step at which the episode terminates. The action-value function $Q_\pi(s_t, a_t) = \mathrm{E}_\pi [G_t|s_t, a_t]$ represents the expected return and is utilized to evaluate the actions selected at the current state under a specific policy. The optimal policy should obtain the optimal action-value function, which has a maximum expected return for each state-action pair:

$$Q_*(s_t, a_t) = \max_\pi Q_\pi(s_t, a_t) \tag{9}$$

The recursive relationships of the optimal action-value function can be formulated with the Bellman optimality equation [30]:

$$Q_*(s_t, a_t) = \mathrm{E}_{s_{t+1}\sim P}[r_t + \gamma \max_a Q_*(s_{t+1}, a)] \tag{10}$$

With this recursive relationship, the optimal action-value function could be estimated with a parameterized function approximator in RL and the parameters can be learned through experience [31]. For example, the agent of Q-learning employs a linear function approximator to estimate the optimal action-value function and derive the optimal policy [32].

### B. Deep Q-learning

In DQL, the agent utilizes a neural network as the function approximator, which could more efficiently approximate the action-value function when the state space of the MDP is huge. In [33], V. Mith et al. proposed a DQL algorithm that is a variant of Q-learning.

In DQL, experience replay is firstly proposed to improve data efficiency and avoid divergence. The transitions ($s_t$, $a_t$, $r_t$, $s_{t+1}$) in experience replay are acquired by an epsilon greedy policy. The DQL algorithm randomly samples $N$ transitions from the experience replay. Then, the samples will be exploited to update the network by minimizing the loss function:

$$L(\theta_t) = \sum_{1}^{N} (y_t^{DQL} - Q(s_t, a_t; \theta_t))^2 \tag{11}$$

wherein the target $y_t^{DQL}$ is derived by the target network $\theta_t^-$ with a greedy policy:

$$y_t^{DQL} = r_t + \gamma \max_a Q(s_{t+1}, a; \theta_t^-) \tag{12}$$

and $\theta_t^-$ is updated by $\theta_t$ only every k time-steps. $\theta_t$ is the prediction network to evaluate the value of the sampled state-action pair. $\theta_t$ will be trained by stochastic gradient descent of the loss function at every time-step:

$$\nabla_{\theta_t} L(\theta_t) = (y_t^{DQL} - Q(s_t, a_t; \theta_t)) \nabla_{\theta_t} Q(s_t, a_t; \theta_t) \tag{13}$$

Although the DQL algorithm has been used in many areas and can usually achieve great performance, there are still drawbacks such as the substantial overestimate.

### C. Double Deep Q-learning

In DQL, the target $y_t^{DQL}$ is calculated by maximizing over the biased estimate from the single network $\theta_t^-$, which can lead to overestimating the action-value function.

Therefore, the double DQL algorithm (named DDQL for short) was proposed to decrease the overestimation by utilizing two networks for the target estimate [34]. In the DDQL, the two networks employed are the prediction network $\theta_t$ and the target network $\theta_t^-$. The prediction network $\theta_t$ is employed to select an action under the greedy policy while the target network $\theta_t^-$ is utilized to evaluate the action. Then, the target $y_t^{DDQL}$ can be derived as:

$$y_t^{DDQL} = (r_t + \gamma Q(s_{t+1}, \arg\max_a Q(s_{t+1}, a; \theta_t); \theta_t^-) \tag{14}$$

After easing the overestimation of the action-value function by DDQL, a better policy can be learned.

### D. Specification of the MDP

In this part, the specifications of states, actions, rewards function and transition model of the MDP for the intersection decision-making problem are proposed.

In this paper, the agent focuses on the positions, velocities and heading angles of the EAV and the surrounding vehicles during the detection area. The state space is defined as:

$$\begin{aligned} & S=(S_e,S_i) \\ & s.t.\, S_e=(p_e,x_e,y_e,v_{x_e},v_{y_e},\sin(\varphi_e),\cos(\varphi_e)) \\ & \quad S_i=(p_i,x_i,y_i,v_{x_i},v_{y_i},\sin(\varphi_i),\cos(\varphi_i)) \\ & \quad s.t.\ p_e,p_i\in\{0,1\} \end{aligned} \tag{15}$$

where $S_e$ is the ego vehicle, $S_i$ is the surrounding vehicle. $p$ indicates whether the vehicle exists that is 1 if the vehicle exists and 0 if it does not. $x$ represents the longitudinal position of the vehicle and $y$ represents the transverse position. $v_x$ is the longitudinal speed and $v_y$ is the lateral speed, $\varphi$ is the heading angle of the vehicle. The detection area is the center of the vehicle itself as a dot, and the detection radius is 60m. The detection radius is calculated by multiplying the maximum vehicle speed by 6 seconds. The velocities of vehicles are limited in the range [0, 10] m/s.

The action controlled by DRL methods is only the longitude acceleration of the EAV. Hence, the discrete action space for longitude acceleration is:

$$a_t\in[-5,\ 0,\ 5]\ \text{m/s}^2 \tag{16}$$

The steering angle $\delta_t$ of the EAV is controlled by the lateral controller proposed in part B of Section II.

As to the surrounding vehicles, their longitude acceleration and steering angle are respectively controlled by the IDM and the lateral controller.

The transition model is built based on a kinematic bicycle model to compute the speed and position changes of each vehicle [35]. The speed $v_{t+1}$ at the center of gravity and the heading angle $\psi_{t+1}$ of one vehicle at the next time-step can be derived with the inputs of acceleration $a_t$ and the steering angle $\delta_t$ as:

$$v_{t+1}=v_t+a_t\cdot dt \tag{17}$$

$$\psi_{t+1}=\psi_t+\omega_t\cdot dt \tag{18}$$

$$\omega_t=v_t\cdot\sin\beta_t\ /\ l \tag{19}$$

$$\beta_t=\arctan(1/2\tan\delta_t) \tag{20}$$

where $dt$ is the length of one time-step, $l$ is the half-length of the vehicle, $\omega_t$ is the yaw rate of the vehicle, $\beta_t$ is the slip angle at the center of the gravity. Then, the velocities $v_{x,t+1}$, $v_{y,t+1}$ at next time-step can be got by breaking the speed of the vehicle in $x$ and $y$ directions:

$$\begin{cases} v_{x,t+1}=v_{t+1}\cdot\sin(\beta_t+\psi_t) \\ v_{y,t+1}=v_{t+1}\cdot\cos(\beta_t+\psi_t) \end{cases} \tag{21}$$

Thus, the positions of the vehicle at the two directions are:

$$\begin{cases} s_{x,t+1}=s_{x,t}+v_{x,t+1}\cdot dt \\ s_{y,t+1}=s_{y,t}+v_{y,t+1}\cdot dt \end{cases} \tag{22}$$

Finally, when the acceleration and steering angle are known,the states at the next time-step of the EAV can be easily acquired by the equations (17-22).

Appropriately setting rewards is significant as rewards direct the learning of the driving policy for the EAV. The reward function is defined as:

$$r=\begin{cases} -5\cdot collision+1\cdot highest\text{-}speed, & \text{if not completing left-turn} \\ 1, & \text{if completing left-turn} \end{cases} \tag{23}$$

where *collision*=1 if the EAV doesn't crash a surrounding car or *collision*=0, and *highest-speed*=1 if the EAV drives at the highest speed or *highest-speed*=0. This reward function is meant to ensure safety and improve efficiency.

## IV. Simulation Results

This decision-making results for the autonomous vehicles are discussed in this section. The training parameters for the DRL methods (e.g., DQL and DDQL algorithms) are first given. Then, the learning processes of these two approaches at the unsignaled intersection are compared. Finally, the trained mature neural networks are verified for a similar driving scenario. The robustness of the obtained decision-making policies is recited.

### A. Comparison Analysis

To conduct a fair comparison between DQL and DDQL, the related parameters in the DRL method are kept the same. The discount factor $\gamma$ is 0.95, and the total number of episodes is 4000. The model of the neural network is a multilayer perceptron with 128 inputs and 3 outputs. The size of the batch is 64, the capacity of the memory is 15000, and the frequency steps to update the target network is 50 [36]. The activation function is a rectified linear unit. The epsilon greedy policy is utilized to choose the control action, the initial value of epsilon is 1, the final epsilon is 0.05, and its value is annealed with 10000 steps.

This subsection compares the training performance of DQL and DDQL in the intersection environment. Fig. 3 depicts the reward curves of these two methods. To describe their differences better, four shapes of the reward are given, which are raw points, original lines, smooth lines and linear fits. The values of reward increase with the number of episodes, which implies that the EAV becomes more familiar with the driving scenario and could adopt more appropriate control actions. In the cases of smooth lines and linear fits, it is easy to see that the DDQL is higher than DQL. This feature means DDQL is capable of achieving better control performance than DQL. Since the high-speed and collision-free targets are included in the reward function (24), DDQL could derive a safer and more efficient decision-making policy.

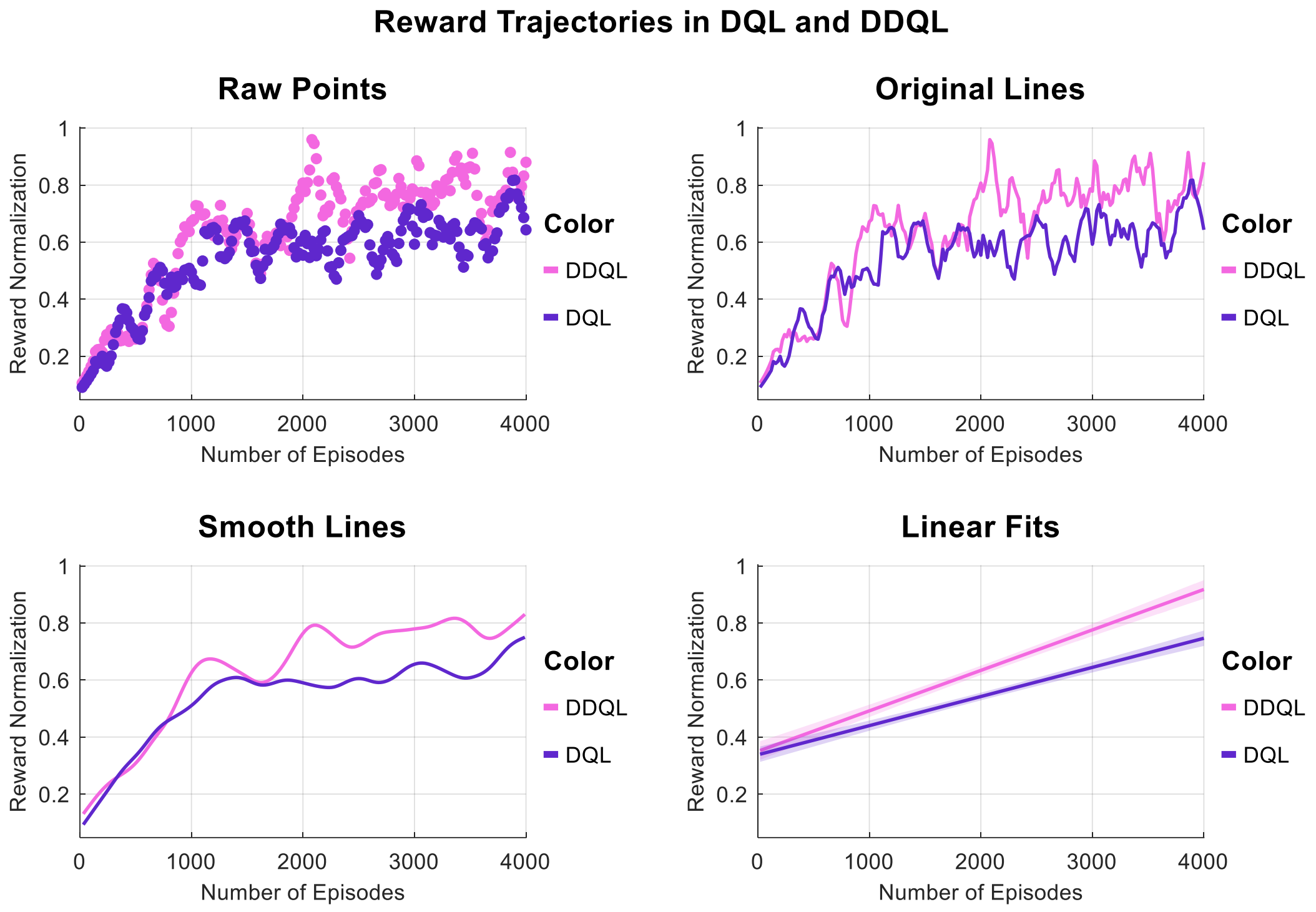

Fig. 3. Normalized reward trajectories in DQL and DDQL with four shapes.

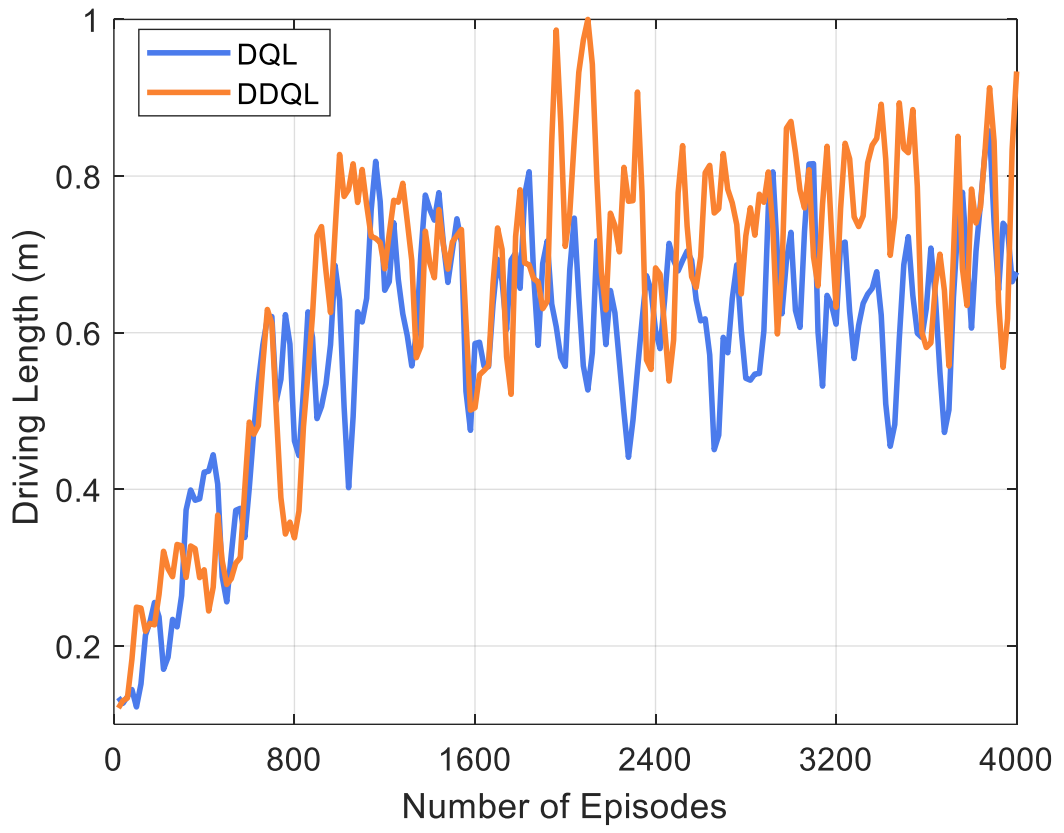

Fig. 4. Driving lengths of each episode for EAV in two DRL methods.

In the training process, the EAV is expected to learn to yield the surrounding vehicles according to the right-of-way rules. The distance between the starting and ending points is fixed, and thus the driving length of each episode could reflect the running efficiency of EAV. The driving lengths in these two DRL techniques are displayed in Fig. 4. Many factors would influence the driving lengths, such as vehicle speed, collision condition, and traveling time. From Fig. 4, the driving length in DDQL is often higher than that in DQL. It indicates that the trained EAV in DDQL is able to go through the intersection with a higher speed and without collision. The fluctuation of the driving length is affected by not only the EAV's own driving strategy but also the traffic conditions of the surrounding vehicles. In the next subsection, the effects of these surrounding vehicles are further analyzed.

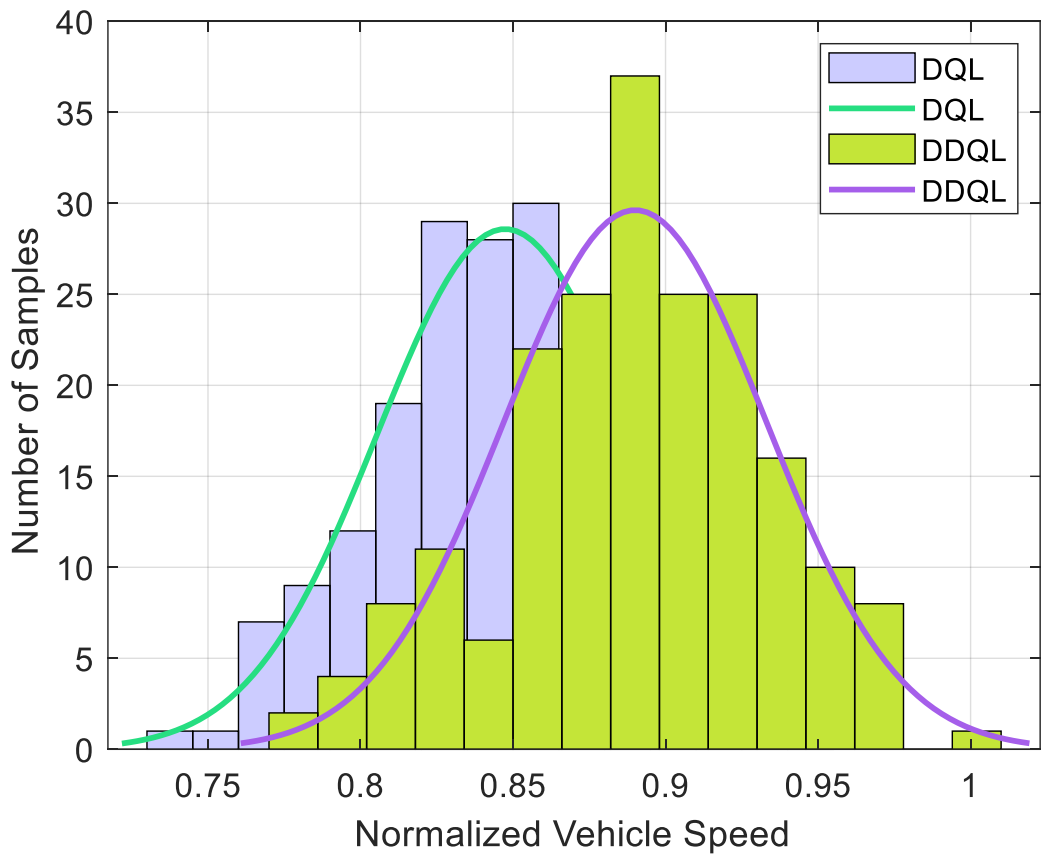

Fig. 5. Vehicle speed distribution of EAV in two compared control cases.

As the vehicle velocity would impact the reward function and the driving length, the vehicle speeds of EAV are highlighted as the frequency histogram in Fig. 5. The X-axis is the value of normalized vehicle speed, and Y-axis is the sample distribution. It can be discerned that the maximum value in DDQL is bigger than that in DQL. The fitted curve further illustrates that the EAV's speeds in DDQL are larger than those in DQL. Hence, the decisions in DDQL could contribute to higher rewards and longer traveling distance. Since the vehicle speed is directly affected by the control actions (acceleration), it can be concluded that the DDQL could realize better control performance in the decision-making problem than DQL. It is owed to the elimination operation of overestimation in DDQL.

Finally, the training efficiency of these two control cases is exhibited. The loss function is the difference between the predicted and actual values in (13). For these two compared methods, multiple types of loss functions are sketched in Fig. 6. To observe these differences better, the raw points, original lines, smooth lines, and linear fits are described in Fig. 6. The original and smooth lines indicate that the loss function decreases with the number of episodes. It means that the action-value function $Q(s, a)$ converges to a mature one based on the collected experiences in the two methods. The lower value of loss function means the higher convergence rate in the DRL approach. From the linear fit case, it is clear that the training speed of DQL is faster than DDQL. This is caused by the two neural networks in DDQL. Therefore, DDQL could achieve better control effectiveness by sacrificing the convergence rate. For the offline requirement, DDQL is suitable to derive the superior decision-making policy.

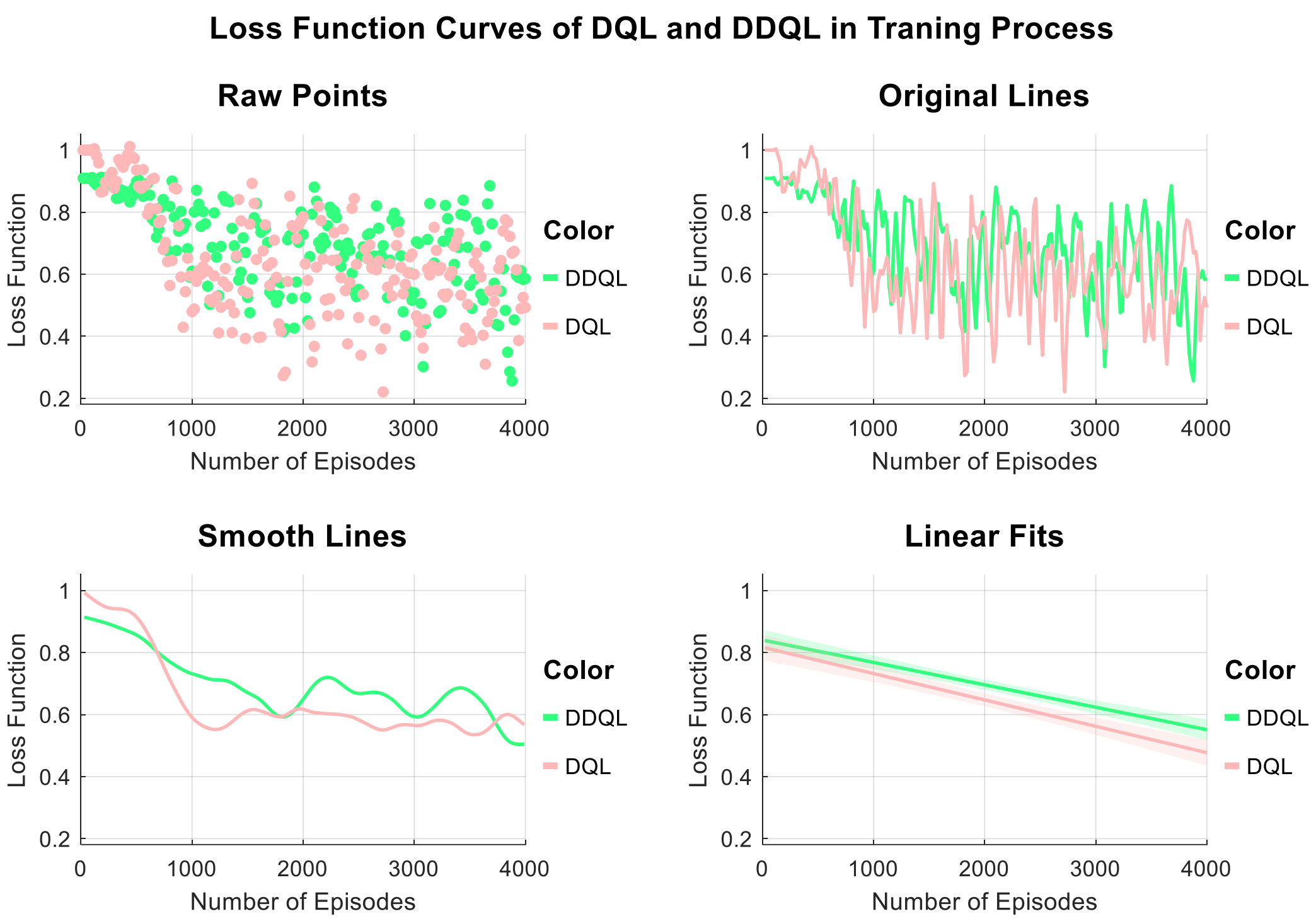

Fig. 6. Convergence rates comparison in DQL and DDQL with four types.

## B. Robustness Evaluation

After training the mature neural network and action-value function, the learned decision-making strategy could be applied in a similar driving scenario. The testing driving environment is the same as the training one, which indicates the left-turn maneuver at the unsignaled intersection. Two policies based on the DQL and DDQL algorithms are estimated. The robustness of these decision-making strategies is concerned. It means whether the EAV can drive through the intersection or not. The number of testing episodes is 40, and the videos of each episode are recorded.

First, the rewards in each testing episode are discussed. For observed convenience, Fig. 7 shows the values of reward in the first 10 episodes. It is easily seen that the rewards in DDQL are a bit larger than those in DQL. The average values of reward in these two approaches are 0.92 and 0.84, respectively. The maximum values are the same, 1. The lowest values in these two control cases are 0.71 and 0.85. It indicates the DDQL based decision-making policy could obtain better control performance than DQL. The collision rates in these two methods are displayed in Table II. It indicates that in most cases, the EAV is capable of running through the intersection without causing a collision. However, some accidents may happen to affect the behaviors of EAV. These traffic conditions will be illuminated in Fig. 9 and Fig. 10.

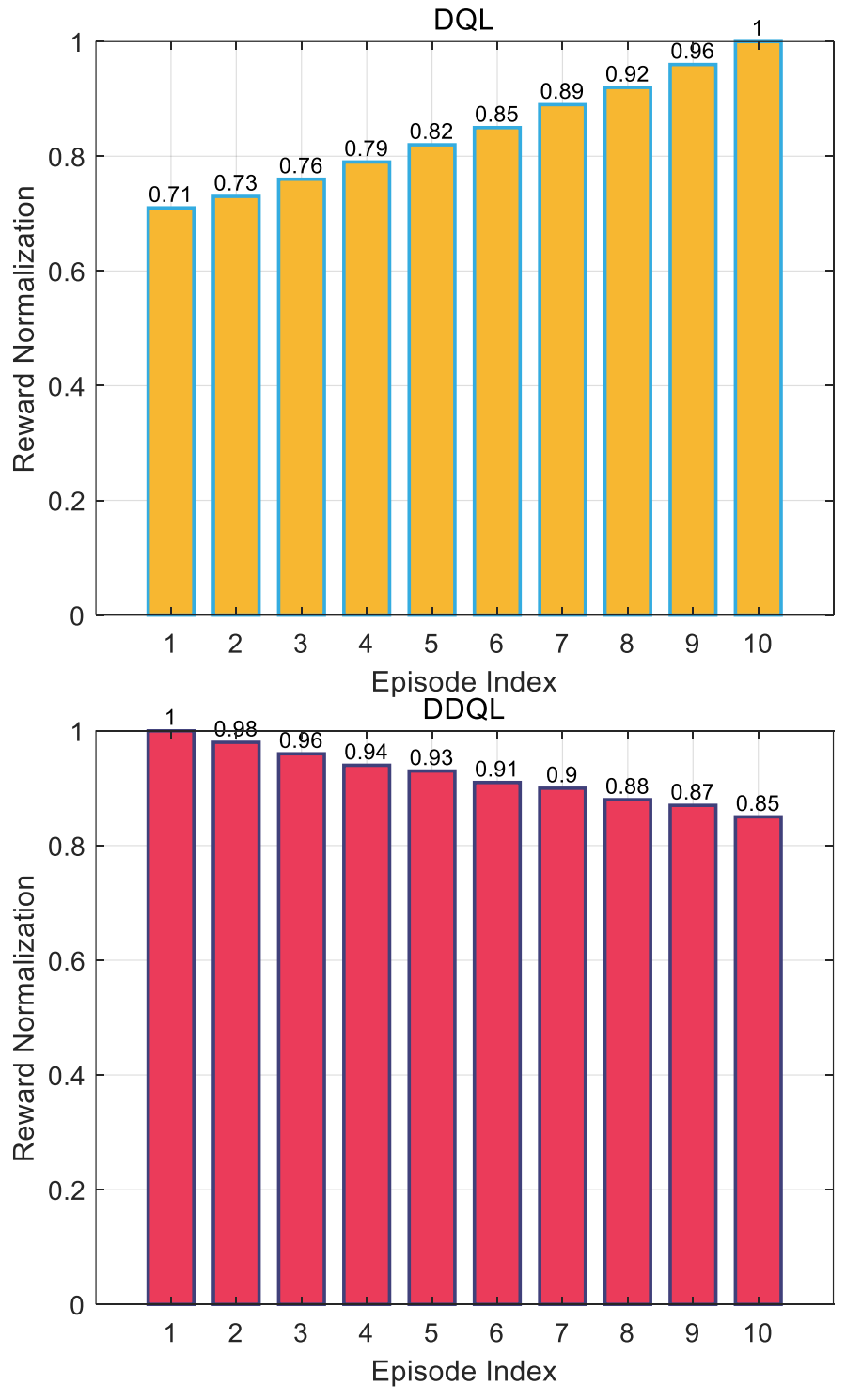

Fig. 7. Values of reward in DQL and DDQL in testing experiments.

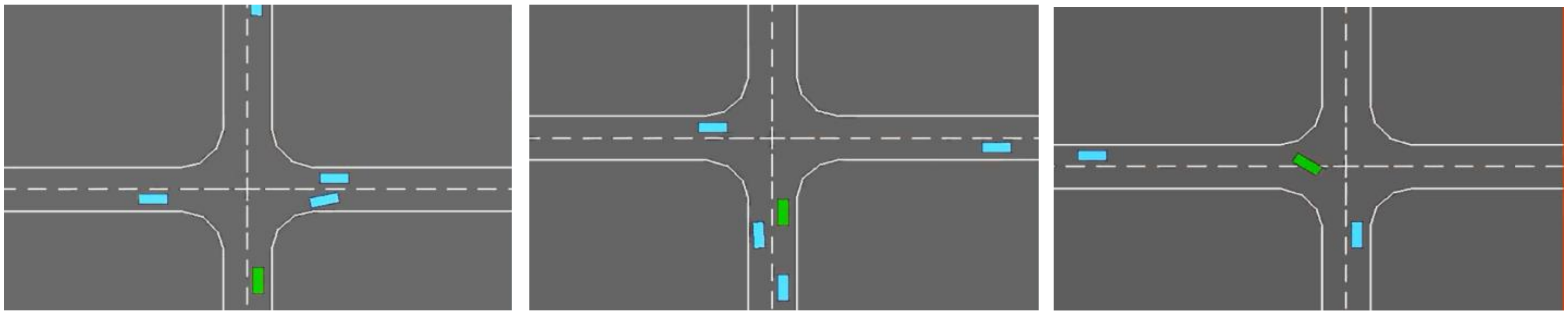

Time Step=2: Arrive the Intersection. Time Step=6: Yield the Surrounding Vehicles. Time Step=10: Execute Left-turn Maneuver.

Fig. 9. Successful left-turn maneuver by yielding the surrounding vehicles.

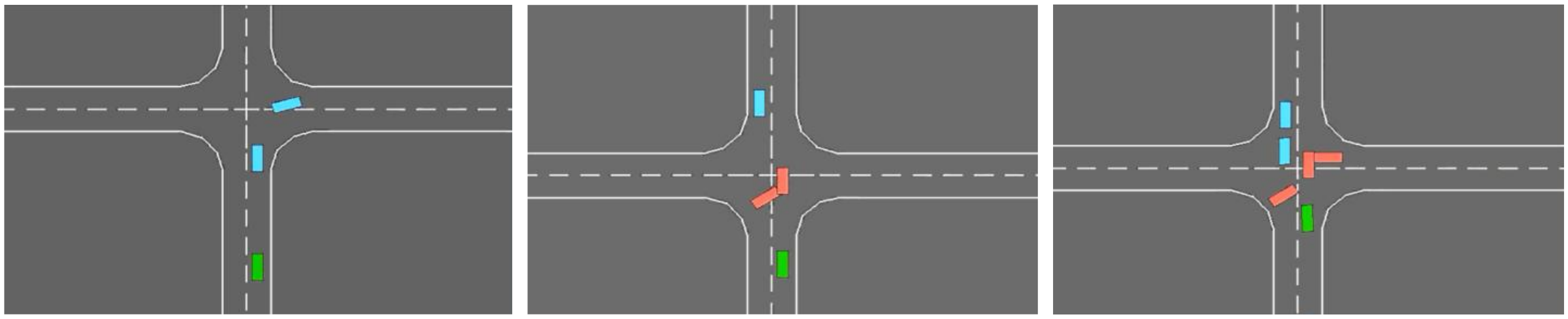

Time Step=1: Arrive the Intersection. Time Step=3: Collision Happens Between Other Vehicles. Time Step=9: EAV Stops to Avoid Collision.

Fig. 10. A failure control case due to the collision between the surrounding vehicles.

TABLE II
COLLISION CONDITIONS IN THREE COMPARED APPROACHES

| Algorithms | Collision rate (%) | Total Episodes |
|---|---|---|
| DQL | 15 | 40 |
| DDQL | 6.25 | 40 |

Since the decision-making policy is constituted by the control actions, the distribution of control actions in these two DRL methods are depicted in Fig. 8. In (17), there are three control actions for EAV to select, and the indexes of these actions are 1, 2, 3, respectively. These actions would influence the vehicle speed directly. In (24), the reward function propels the EAV to run as fast as possible. Thus, the bigger index of control actions means higher vehicle speed. From Fig. 8, it is obvious that the maximum and median in DDQL are larger than those in DQL. This characteristic implies that the control actions of DDQL are better than DQL in the decision-making problem of this work. The robustness and adaptability of DDQL-enabled decision-making policy are preferable.

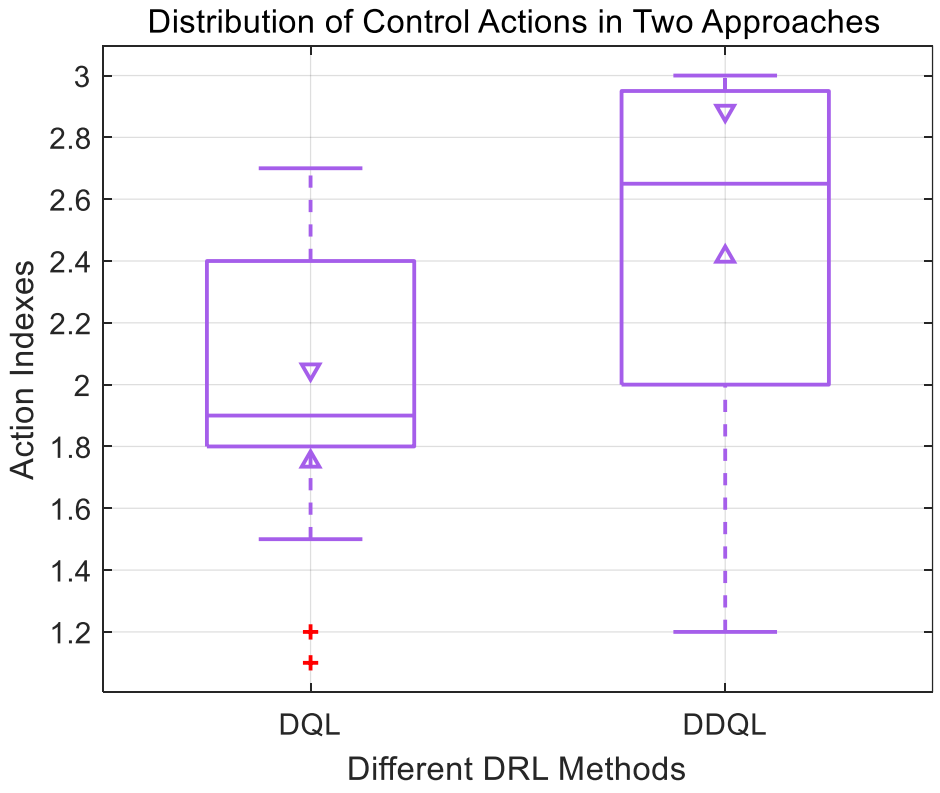

Fig. 8. Box distribution of control action indexes in DQL and DDQL.

Finally, two typical traffic conditions are given in Fig. 9 and Fig. 10 by the recording videos of the DDQL algorithm. In Fig. 9, the EAV is successful in going through the intersection. Based on the frames at different time steps, it can be seen that the EAV has learned to yield other vehicles because of the lowest right-of-way. After other surrounding vehicles driving through the intersection, the EAV executed a left-turn maneuver. Furthermore, Fig. 10 states one accident of the surrounding vehicles. Since the speed and position of these vehicles are randomly given, and the behaviors of these vehicles are managed by IDM, a collision may happen between two surrounding vehicles. Then, the EAV recognizes that a collision may happen when it goes through the intersection. Hence, the EAV stops to avoid a collision. This situation would affect the driving length in Fig. 4 and the reward in Fig. 7. It also indicates that the DRL method is superior to the baseline IDM. In conclusion, the proposed decision-making policy could guide the EAV to drive safely and efficiently at the unsignaled intersection. DDQL is more appropriate for this decision-making problem than DQL.

## V. CONCLUSION

In this work, the decision-making problem at the unsignaled intersection is discussed. The special driving behavior of the EAV is the left-turn maneuver. Two DRL methods, DQL and DDQL are utilized to resolve this problem. The training process and evaluated experiments are compared to demonstrate the DDQL is more suitable for this decision-making problem than DQL. However, the convergence rate of DDQL is slower than DQL, which is not an important concern in offline training.

Future work focuses on the online application of the proposed decision-making policy. At that point, more advanced DRL algorithms may be used. Moreover, since the behaviors of the surrounding vehicle will affect the EAV's decisions, the connected environment may be discussed. Finally, the DRL-based decision-making policy can be estimated by the real-world collected driving data.

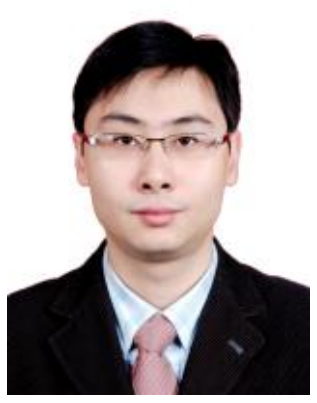

**Feng Wang** received the Ph.D. degree from Chongqing University, China, in 2007. He is currently a Professor and Head of Department of Applied Economics at School of Economics and Business Administration, Chongqing University. He also a research fellow at Institution of Energy Economics, Chongqing University. He was a Visiting Scholar with the Charles H. Dyson School of Applied Economics and Management, Cornell University, Ithaca, NY, USA, between 2013 and 2014, as well as at the Nanyang Business School and National Intuition of Education at Nanyang Technological University, Singapore, in 2013 and 2017. His research interesting includes Traffic Control, Energy System Optimization and Technological Economics.

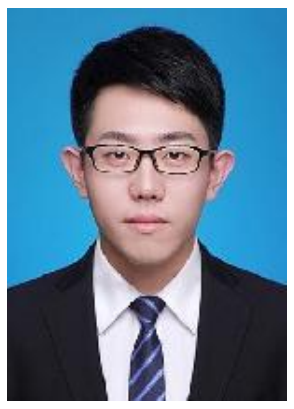

**Dongjie Shi** received the B.S degree in Chongqing University, major in Energy Economics. He is currently pursuing Ph.D. degree in Chongqing University, major in technical economy and management. His current research focuses on the deep reinforcement learning and decision-making for autonomous driving.

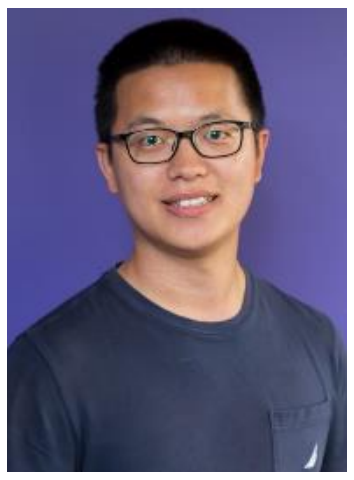

**Teng Liu** (M'2018) received the B.S. degree in mathematics from Beijing Institute of Technology, Beijing, China, 2011. He received his Ph.D. degree in automotive engineering from Beijing Institute of Technology (BIT), Beijing, in 2017. His Ph.D. dissertation, under the supervision of Pro. Fengchun Sun, was entitled "Reinforcement learning-based energy management for hybrid electric vehicles." He worked as a research fellow in Vehicle Intelligence Pioneers Ltd from 2017 to 2018. Dr. Liu worked as a postdoctoral fellow at the Department of Mechanical and Mechatronics Engineering, University of Waterloo, Ontario N2L3G1, Canada from 2018 to 2020. Now, he is a member of IEEE VTS, IEEE ITS, IEEE IES, IEEE TEC, and IEEE/CAA.

Dr. Liu is now a Professor at the Department of Automotive Engineering, Chongqing University, Chongqing 400044, China. He has more than 8 years' research and working experience in renewable vehicle and connected autonomous vehicle. His current research focuses on reinforcement learning (RL)-based energy management in hybrid electric vehicles, RL-based decision making for autonomous vehicles, and CPSS-based parallel driving. He has published over 40 SCI papers and 15 conference papers in these areas. He received the Merit Student of Beijing in 2011, the Teli Xu Scholarship (Highest Honor) of Beijing Institute of Technology in 2015, "Top 10" in 2018 IEEE VTS Motor Vehicle Challenge and sole outstanding winner in 2018 ABB Intelligent Technology Competition. Dr. Liu is a workshop co-chair in 2018 IEEE Intelligent Vehicles Symposium (IV 2018) and has been reviewers in multiple SCI journals, selectively including IEEE Trans. Industrial Electronics, IEEE Trans. on Intelligent Vehicles, IEEE Trans. Intelligent Transportation Systems, IEEE Transactions on Systems, Man, and Cybernetics: Systems, IEEE Transactions on Industrial Informatics, Advances in Mechanical Engineering.

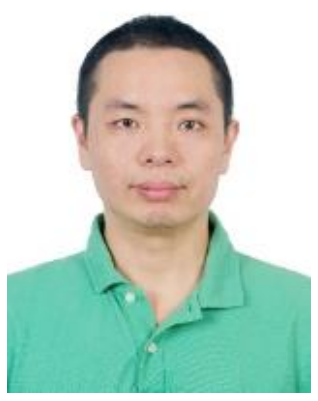

**Xiaolin Tang** received a B.S. in mechanics engineering and an M.S. in vehicle engineering from Chongqing University, China, in 2006 and 2009, respectively. He received a Ph.D. in mechanical engineering from Shanghai Jiao Tong University, China, in 2015. From August 2017 to August 2018, he was a visiting professor of Department of Mechanical and Mechatronics Engineering, University of Waterloo, Waterloo, ON, Canada. He is currently an Associate Professor at the College of Mechanical and Vehicle Engineering, Chongqing University, Chongqing, China. He is also a committeeman of Technical Committee on Vehicle Control and Intelligence of Chinese Association of Automation (CAA). He has led and has been involved in more than 10 research projects, such as National Natural Science Foundation of China, and has published more than 30 papers. His research focuses on Hybrid Electric Vehicles (HEVs), vehicle dynamics, noise and vibration, and transmission control.